\documentclass[10pt,twocolumn,letterpaper]{article}

\usepackage{cvpr}              

\usepackage{graphicx}
\usepackage{amsmath}
\usepackage{amssymb}
\usepackage{booktabs}
\DeclareMathOperator{\E}{\mathbb{E}}

\usepackage[pagebackref,breaklinks,colorlinks]{hyperref}

\usepackage[capitalize]{cleveref}
\crefname{section}{Sec.}{Secs.}
\Crefname{section}{Section}{Sections}
\Crefname{table}{Table}{Tables}
\crefname{table}{Tab.}{Tabs.}

\def\cvprPaperID{12054} 
\def\confName{CVPR}
\def\confYear{2022}

\begin{document}

\title{Face Mask Removal with Region-attentive Face Inpainting}
\author{Minmin Yang\\
{\tt\small \{myang47\}@syr.edu }}
\maketitle

\begin{abstract}
During the COVID-19 pandemic, face masks have become ubiquitous in our lives. Face masks can cause some face recognition models to fail, since they cover significant portion of a face. In addition, removing face masks from captured images or videos can be desirable, \eg, for better social interaction and for image/video editing and enhancement purposes. Hence, we propose a generative face inpainting method to effectively recover/reconstruct the masked part of a face. Face inpainting is more challenging compared to traditional inpainting, since it requires high fidelity while maintaining the identity at the same time. Our proposed method includes a Multi-scale Channel-Spatial Attention Module (M-CSAM) to mitigate the spatial information loss and learn the inter- and intra-channel correlation. In addition, we introduce an approach enforcing the supervised signal to focus on masked regions instead of the whole image. We also synthesize our own Masked-Faces dataset from the CelebA dataset by incorporating five different types of face masks, including surgical mask, regular mask and scarves, which also cover the neck area. The experimental results show that our proposed method outperforms different baselines in terms of structural similarity index measure, peak signal-to-noise ratio and $\ell_1$ loss, while also providing better outputs qualitatively. The code will be made publicly available. Code is available at \href{https://github.com/LexieYang/Face_Img_Inpainting}{GitHub}.

\end{abstract}

%
\label{sec:intro}

\section{Introduction}
Image inpainting is a process to remove undesired parts and/or restore corrupted regions of images by filling in these areas. One of the challenges with image inpainting is how to reasonably utilize information from known regions to synthesize realistic and occlusion-free images while maintaining structure and texture consistency. 

Compared to inpainting of outdoor images, face inpainting is more challeging, since face completion requires more attention to topological anatomy and persistent attributes. Face masks have been playing an important role in protecting lives, especially during the COVID-19 pandemic. However, face masks can also degrade the performance of face recognition systems. 
Zhu \etal~\cite{zhu2021masked} report that the performance of the ArcFace~\cite{deng2019arcface} face recognition model drops significantly on the masked facial images.

Traditional works can be broadly classified into two categories: patch-based and diffusion-based approaches.~Patch-based approaches~\cite{criminisi2004region,guo2017patch} fill the unknown region by utilizing the redundancy of the image itself. These methods aim to find the best matching block, for the patch to be filled, in the known area and copy it to the corresponding target area. Diffusion-based approaches~\cite{telea2004image, bertalmio2000image} start restoring from the boundary of the unknown region. All pixels in the region are gradually filled from the boundary to the center, and the pixels to be filled are obtained by the weighted sum of all known pixels in the neighborhood.
Even though these achievements have elucidated some of the underlying challenges with image inpainting, they have limited performance when the missing area has a complex texture and structure.

Rapid development of deep convolutional neural networks (CNNs) and generative adversarial networks (GANs) has motivated their use for image inpainting~\cite{hui2020image, fang2020face, iizuka2017globally}. CNN-based methods can achieve better results by learning rich semantic information from large-scale datasets and repairing the missing regions in an end-to-end way. GANs are incorporated for better structural coherency and content rationality. Results of these works show that they can generate plausible content with consistent texture.

Using the contextual information around unknown pixels is the key idea in ~\cite{wang2019musical, Yu_2019_GC, Yan_2018}. They only focus on semantic consistency between the generated pixels and known pixels. Liu \etal~\cite{Liu_2019}, on the other hand, further consider the consistency inside the corrupted area, by proposing a two-step strategy. In the first step, the known pixels are used to initialize unknown pixels for semantic consistency between the generated pixels and ground-truth pixels. In the second step, the results are optimized by considering adjacent consistency inside the unknown area for local feature coherency. The aforementioned methods rely on cosine similarity to measure the similarity between patches. However, they only take the spatial information into consideration when calculating the attention map.

In previous works~\cite{pathak2016context,li2017generative, yu2020region, shin2020pepsi++}, an encoder-decoder architecture with vanilla convolutions is adopted to remove unwanted areas while generating desired contents. The encoder encodes the input image with masked region and extracts latent feature representation of the image. The decoder takes the hidden representation as input and outputs a recovered image. However, due to the specificity of image inpainting task that masked pixels in the image are not valid, encoder-decoder with vanilla convolutions does not work well. The vanilla convolution treats all pixels (masked and unmasked) equally, which may lead to visual artifacts, like color discrepancy and structure distortion. To overcome this limitation, GatedConv~\cite{yu2019free} is proposed, which formulates mask updates as soft-gating. 

In this work, we propose a Multi-scale Channel-Spatial Attention Module (M-CSAM), which applies a channel-spatial attention mechanism~\cite{niu2020single} to guide the restoration. The proposed M-CSAM incorporates the spatial and channel information to learn the inter-channel and intra-channel correlation simultaneously.

\begin{figure*}[ht]
    \centering
    \includegraphics[width=20cm]{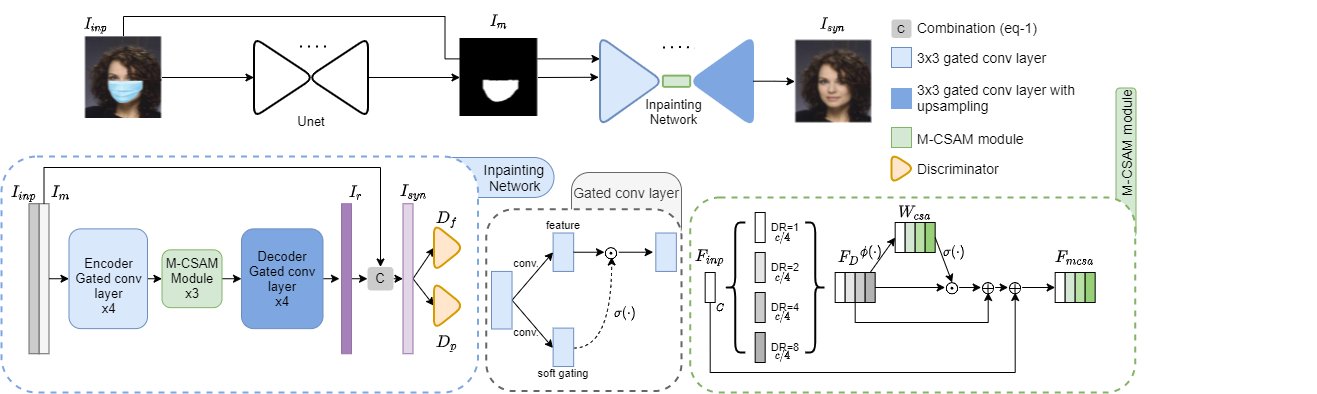}
    \caption{\textbf{The main architecture of the proposed approach.} The legend shows what each colored box refers to.}
    \label{fig:NetArch}
\end{figure*}

To improve the practicality of our approach, and to remove the assumption that masked regions are known, we first adopt a modified version of U-Net~\cite{long2015fully} to segment out the masked region in a masked face image. The overall workflow for mask segmentation and face image restoration is showed in \cref{fig:NetArch}. Experiments show that the Intersection over Union (IoU) between a generated mask image and the ground-truth mask image is over $0.99$ showing that the segmented mask area can be useful to assist in the recovery/inpainting of the masked face image.

After segmenting the face mask areas, we propose an encoder-decoder structure with gated convolution layers~\cite{yu2019free} to fill the masked region gradually and adaptively. In contrast to vanilla convolution, which treats every pixel the same way, gated convolution can dynamically select features channel-wise and spatial-wise. In the encoder, the down-sampling operation helps to reduce the spatial dimension while extracting more features. With the utilization of skip-connections between the encoder and decoder, the decoder can restore fine-grained information from the down-sampling path and recover the image with more consistent texture and structure. Between the encoder and decoder, three M-CSAMs are adopted to efficiently learn the texture and structure feature maps. The M-CSAM is composed of two steps. First, it applies dilated convolution with four dilation rates (1,2,4 and 8) to the input feature map. After the dilated convolutions, these four feature maps are concatenated along the spatial dimension to be further processed by the channel-spatial attention module (CSAM)~\cite{niu2020single}. The CSAM~\cite{niu2020single} incorporates $3\times3\times3$ 3D convolutions to extract inter-channel and intra-channel correlation, resulting in an attention map. Then, the attention map is multiplied with the input of the CSAM. The attention feature map is scaled by a factor and added to the input of the CSAM to get the output. Finally, the output of CSAM is added to the input of M-CSAM to get the output of M-CSAM. 

\textbf{Contributions.} The main contributions of this work include the following: 
\begin{itemize}
\item We propose a Multi-scale Channel-Spatial Attention Module (M-CSAM) that incorporates a spatial pyramid structure with Channel-Spatial Attention Module to increase the representativeness of the network.
\item We propose to use region-attentive supervision by enforcing the supervised signal to focus only on the masked area of the face instead of the whole image to limit the variance of generated content.

\item The proposed approach outperforms four state-of-the-art (SOTA) baselines in terms of structural similarity index measure, peak signal-to-noise ratio and $\ell_1$ loss, while also providing better outputs qualitatively. 

\item We synthesize a new face mask dataset, named Masked-Faces, from the CelebA dataset by incorporating five different types of masks, which include surgical masks, regular masks and scarves.
\end{itemize}


\label{sec:format}
\section{Related Work}
\subsection{Image inpainting}
Traditional diffusion-based and patch-based methods utilize local or non-local information to synthesize the missing contents based on single image~\cite{bertalmio2000image, levin2003learning, ballester2001filling,sun2005image, barnes2009patchmatch, huang2014image}. 
Images are rotated and scaled iteratively to obtain the matching patches by patch matching (PM) method~\cite{mansfield2011transforming, kai2015image}. Although these methods can solve the problem of rotation and scaling to some extent, they are time-consuming and memory-intensive. Huang \etal~\cite{huang2014image} propose a plane structure guidance (PSG) for image restoration. Levin \etal~\cite{levin2003learning} attempt to tackle the global consistency problem by statistical learning. They restore the hole/missing region by building a family distribution over images and searching for the best matching image. 

Recently, deep learning- and GAN-based methods~\cite{hui2020image, fang2020face, yu2020region, iizuka2017globally} have been shown to be effective to generate plausible images visually and structurally. The method of adopting local and global discriminators to ensure global and local content consistency is widely used~\cite{hui2020image, li2017generative, li2019face}. Instead of generating one result for each occluded image, Zheng \etal~\cite{zheng2019pluralistic} present a probabilistic framework to generate multiple and diverse images. For image inpainting, the pixels in the masked region are viewed as invalid. However, CNN-based methods, which use standard convolutions, treat masked pixels same way as the valid ones. This leads to generating blurry images with distorted structure. Liu \etal~\cite{liu2018image} propose partial convolutions that are conditioned only on valid pixels. They also include a rule-based method to automatically update mask for next layer. Partial convolution classifies pixels as invalid or valid, and the one-or-zero masks are multiplied with inputs for all layers. However, heuristically categorizing all pixels as invalid or valid may cause severe information loss. Therefore, Yu \etal~\cite{yu2019free} design gated convolution for free-form image inpainting.
The mask is viewed as a soft-gated learnable channel and updated automatically from data. Yu \etal~\cite{yu2020region} propose a region normalization (RN) method that computes the mean and variance in masked and unmasked region, respectively, for normalization.

\subsection{Face Completion}
Face completion requires complex and delicate texture generation with large variation. In addition, face completion requires high semantic and contextual coherency, since each facial component is closely related. Li \etal~\cite{li2017generative} propose a generator and two discriminators to generate plausible contents from random noise. In addition, a semantic parsing network is designed for regularizing the generation process. This model is sensitive to pose/orientation of the face, and cannot perform well when face images are not well-aligned. Song \etal~\cite{song2019geometry} propose a facial geometry estimator to predict facial landmark heatmaps and parsing maps. Then, the estimated facial geometry images and masked face images are sent to the encoder-decoder structured generator for image completion. The estimated facial geometry information can be used to modify face attributes and perform user-interactive face completion. Instead of proposing an additional network to assist in the generation of high-frequency face details, Wang \etal~\cite{wang2020recurrent} take full advantage of the multi-level features for high-frequency detail generation via two deep CNNs and a recurrent neural network (RNN). The first CNN learns multi-level features, and the RNN transfers the features into another domain. The second CNN takes advantage of the multi-level transferred features and reconstructs the face image.

\subsection{Attention-based image inpainting}
Attention is widely used in different computer vision tasks, such as classification~\cite{residualattention2017, Hu_2018_SENET, guo2019visual, Wang_2018_non_local, zhu2020learning}, segmentation~\cite{tao2020hierarchical, fu2019dual, huang2019ccnet, yuan2019objectcontextual, Lu_2019} and image generation~\cite{gou2020segattngan, Xu_2018, gregor2015draw, kalarot2020component}. For the image inpainting task, Yu \etal~\cite{Yu_2018_CA} propose a two-stage network, predicting a coarse result at the first stage and refining it at the second stage. The contextual attention (CA) layer used at the second stage calculates the similarity between unknown patches and background patches based on the previous coarse prediction. Liu \etal~\cite{Liu_2019} propose a coherent semantic attention layer (CSA), which not only calculates the similarity between known and unknown patches, but also the similarity among the unknown patches. Wang \etal~\cite{wang2019musical} propose a multi-scale image contextual attention learning (MUSICAL) strategy, which applies two different patch sizes ($3 \times 3 $ and pixel-wise) and generates two feature maps. Then, two feature maps are concatenated and re-weighted by a Squeeze-and-Excitation module~\cite{hu2018squeeze}. Different from the previous contextual attention mechanisms, Hui \etal~\cite{hui2020image} propose a self-guided map, which could also be viewed as spatial attention map. They also introduce a self-guided regression loss to measure the hard position in the feature space. The original CA~\cite{Yu_2018_CA} computes the similarity between known and unknown regions, while CSA~\cite{Liu_2019} expands it to keep the inside of the unknown region coherent.

\label{sec:method}

\section{Proposed Method}\label{sec:proposed}
The main architecture of the proposed method is presented in \cref{fig:NetArch}. It consists of two primary networks: a segmentation network, which is a modified version of U-Net~\cite{ronneberger2015u}, and an inpainting network. In the first stage, the segmentation network segments out the mask region to generate a binary mask image. In the second stage, the masked face image and the binary mask image from the first stage are sent together to the inpainting network to complete the missing facial structures.

\subsection{Segmentation Network}\label{ssec:Segmentation}
Given an image with a face mask, the objective of the segmentation network is to generate a binary segmentation map of mask, $I_m$, wherein $1$ and $0$ represent the masked and unmasked region, respectively. We construct a modified version of U-Net~\cite{long2015fully} to generate the binary segmentation mask. The segmentation network consists of an encoder and a decoder with skip connections via concatenation. The encoder consists of four blocks for down-sampling. Each encoder block includes a max-pooling layer, and two convolutional blocks. The convolutional block has a convolution layer followed by a Batch Normalization layer and a ReLU activation function. The decoder consists of 4 decoder blocks for up-sampling. Each decoder block is composed of a bilinear interpolation layer for up-sampling, and also two convolutional blocks. The last layer of the decoder uses the \textit{sigmoid} activation function. Binary cross-entropy loss is used as the objective function. 

\subsection{Inpainting Network} \label{ssec:Inpainting}
We adopt an encoder-decoder structure with gated convolution~\cite{yu2019free} to recover the masked area of faces via image inpainting. The encoder-decoder with the skip connections is a modified version of the generative network~\cite{isola2017image}.  Let $I_{inp}$, $I_{m}$, $I_{r}$ and $I_{gt}$ denote the input image, the binary mask image, the output of the generator and the ground truth, respectively. Since our objective is to recover the masked region, we only consider the loss of the restoration results in the masked region, and combine the unmasked region from the input with the recovered region from the generator's output, as shown at the bottom left of \cref{fig:NetArch}. The synthetic generated image is denoted by  $I_{syn}$, where
\begin{equation}
    I_{syn}=(1 - I_m)\times I_{inp} + I_m \times I_{r}.
\end{equation}
The input image $I_{inp}$ and the binary mask image $I_m$ are concatenated along the channel dimension and fed to the network. Then, two discriminators, namely a patch discriminator and a feature patch discriminator~\cite{liu2019coherent}, are utilized to enforce an output with high visual quality, and consistency, which are measured by the structural similarity index measure and peak signal-to-noise ratio. \vspace{-0.3cm}


%
\subsubsection{Gated Encoder-Decoder}
There are four gated convolution layers in the encoder as well as in the decoder, with four skip connections between the encoder and the decoder (as shown in \cref{fig:NetArch}). 

Gated convolution proposed by Yu \etal~\cite{yu2019free} is a convolution layer that integrates soft-gating mechanism for image inpainting. It overcomes the limitations of vanilla convolution, which treats all the pixels the same way. Gated convolution  generalizes partial convolution by learning a dynamic feature selection method for every channel at every spatial position. Considering the different shapes of face masks, we incorporate the gated convolution in the encoder and the decoder to learn the masks autonomously. 

\subsubsection{Multi-scale Channel-Spatial Attention Module}
\label{MCSAM}
We propose and incorporate three Multi-scale Channel-Spatial Attention Modules (M-CSAM), which are attention-based dilated residual blocks, between the encoder and the decoder. The M-CSAM has a spatial pyramid structure that evaluates the features with different receptive fields, as shown at the bottom right of \cref{fig:NetArch}. We concatenate the feature maps at different receptive scales, and pass them to an attention module, which is based on Channel-Spatial Attention Mechanism (CSAM)~\cite{niu2020single}, to improve the feature representation ability in multi-scale context.~CSAM~\cite{niu2020single} considers correlations at all the positions of each channel, and is effective in capturing informative features while discarding unnecessary information. \vspace{-0.2cm}

\paragraph{Review of Channel-Spatial Attention Module.}
\label{csam}
Instead of only focusing on the features' spatial dimension 
or proposing channel attention techniques neglecting scale information, a novel channel-spatial attention module (CSAM) is proposed in~\cite{niu2020single}, which takes all feature map dimensions into consideration, including spatial and channel dimensions.~Compared to traditional spatial-channel attention, CSAM adaptively learns inter-channel and intra-channel feature representations.

The architecture of CSAM is shown at the bottom right of \cref{fig:NetArch}. Let $F_D \in R^{H\times W \times C}$ be the feature map that is generated by concatenating feature maps with four dilation rates.~$F_D$ is fed to a 3D convolutional layer~\cite{ji20123d} with kernel size of $3 \times 3 \times 3$ and stride of 1. The 3D kernels are convolved with the input feature maps on consecutive channels, modeling inter-channel and intra-channel dependencies.~Convolution with a set of 3D kernels results in channel-spatial attention feature map $W_{csa}$. Then, the attention map $W_{csa}$ is multiplied with the input feature maps $F_D$. Finally, the weighted feature map is scaled by a factor $\beta$ and added to $F_D$. This process can be shown as: 
\begin{equation}
     W_{csa} = \phi(F_D)
\end{equation}
\begin{equation}
    F_{csa} = \beta \sigma (W_{csa})\odot F_D + F_D,
\end{equation}
where $\phi(\cdot)$ is the 3D convolution operation, $\sigma(\cdot)$ is the \textit{sigmoid} function, $\odot$ is the element-wise multiplication and $F_{csa}$ is the weighted feature map. 

\paragraph{Multi-scale CSAM (M-CSAM).}
In the encoder, the feature size is reduced after each convolutional layer, and some spatial information may be lost. To mitigate the spatial information loss, we perform multi-scale filling by dilated convolutions with different dilation rates for multi-scale response. As shown at the bottom right of \cref{fig:NetArch}, we perform four parallel dilated convolutions with the dilation rates (DR) of 1, 2, 4 and 8. Let $F_{inp}$ denote the $C$-channel input feature map for M-CSAM, and $F^{1},F^{2},F^{4}$ and $F^{8}$ represent the outputs of four dilated convolutional layers (each with $C/4$ channels), respectively. The outputs of the dilated convolutions, obtained with four different dilation rates, can be easily aggregated thanks to output features being the same size. The outputs are concatenated along the channel dimension to produce a multi-scale feature response, $F_D$. 
\begin{equation}
    F_D = [F^1, F^2, F^4, F^8],
\end{equation}
where $[\cdot]$ denotes the concatenation operation.

Then, the feature map $F_D$ is fed to the channel-spatial attention module to calculate the weighted feature map $F_{csa}$. Finally, in order to alleviate the gradient vanishing issue, and improve the gradient flow, we adopt a residual connection for the output of the M-CSAM, which is denoted by $F_{mcsa}$.
\begin{equation}
    F_{mcsa} = F_{inp}+F_{csa}.
\end{equation}
CSAM~\cite{niu2020single} has been demonstrated to be effective in improving the structure and details of images in the single image super-resolution task. Therefore, we adopted CSAM~\cite{niu2020single} and combined it with multi-scale feature maps to effectively learn feature representations and correlation. \vspace{-0.4cm}

\paragraph{Discriminators.}
Instead of adding a local discriminator, we adopt the feature patch discriminator~\cite{liu2019coherent} and $70\times70$ patch discriminator in our model. The feature patch discriminator discriminates whether the input is real or fake by inspecting the feature maps. $70\times70$ patch discriminator, on the other hand, discriminates between the generated and original images by inspecting their pixel values. With the assistance of these two discriminators, our network trains faster, more stable, and can synthesize more meaningful high-frequency details. For the discriminators, we adopt the Relativistic Average LS adversarial loss~\cite{jolicoeur2018relativistic}, which can further stabilize the training process. The loss functions for the generator and the discriminator are denoted by  $L_{gen}$ and $L_{disc}$, respectively, which are defined as:
\begin{equation}
\begin{aligned}
    L_{gen} = -\E_{I_{gt}}[D(I_{gt},I_{syn})^2]-\E_{I_{syn}}[(1-D(I_{syn},I_{gt}))^2] \\
    L_{disc} = -\E_{I_{gt}}[(1-D(I_{gt},I_{syn}))^2]-\E_{I_{syn}}[D(I_{syn},I_{gt})^2] 
    \end{aligned}
\end{equation}
where D represents the discriminator, and $\E_{I_{gt}/I_{syn}}$ stands for the averaging operation.
\vspace{-0.3cm}
\paragraph{Objectives of the inpainting network.}
\label{obj}
For better restoration results, we adopt several loss functions during training, including Relativistic Average LS adversarial loss~\cite{jolicoeur2018relativistic}, reconstruction loss, perceptual loss and style loss.

All the loss functions are computed between a synthetic image and ground truth. Our strategy of 
supervision over the masked region, instead of the whole image, provides better performance. The reason is that since the variance of the masked region is less than the variance of the whole image, it is easier for the network to learn to generate consistent color and structure for the masked region. Ablation studies have been performed to show that local supervision achieves better performance.

The reconstruction loss calculates pixel-wise difference between the network prediction and the ground truth, using the $L_1$ distance, as follows:
\begin{equation}
    L_{r}=||I_{syn}-I_{gt}||_1.
\end{equation}

The perceptual loss considers the correlation between the composite predictions and the original images at feature level to capture high-level semantics. It is formalized as follows:
\begin{equation}
    L_{p} = \sum_{i}\frac{1}{W_i\times H_i\times C_i}||F_i^{syn}-F_i^{gt}||_2
\end{equation}
where $F_i^{syn}$ and $F_i^{gt}$ are the activation maps at the $i$-th layer of VGG-16 backbone with size of $W_i\times H_i\times C_i$. 

It has been demonstrated that the style loss is able to remove checkerboard artifacts \cite{nazeri2019edgeconnect}. It is defined as: 
\begin{equation}
L_s = \sum_{i}\frac{1}{C_i\times C_i}||G_i^{syn} - G_i^{gt}||_1,
\end{equation}
where $G_i^{syn}$ and $G_i^{gt}$ are Gram matrices from the $i$-th selected activation maps, which are the same activation maps used for calculating the perceptual loss.

Combining the reconstruction loss, the perceptual loss, the style loss and the adversarial loss, the overall loss function is defined as follows:
\begin{equation}
    L = \lambda_rL_r + \lambda_pL_p + \lambda_sL_s + \lambda_{adv}L_{adv},
\end{equation}
where $\lambda_r$, $\lambda_p$, $\lambda_s$ and $\lambda_{adv}$ are the corresponding loss weights.

\begin{figure}[t]
\begin{center}
  \includegraphics[width=1\linewidth]{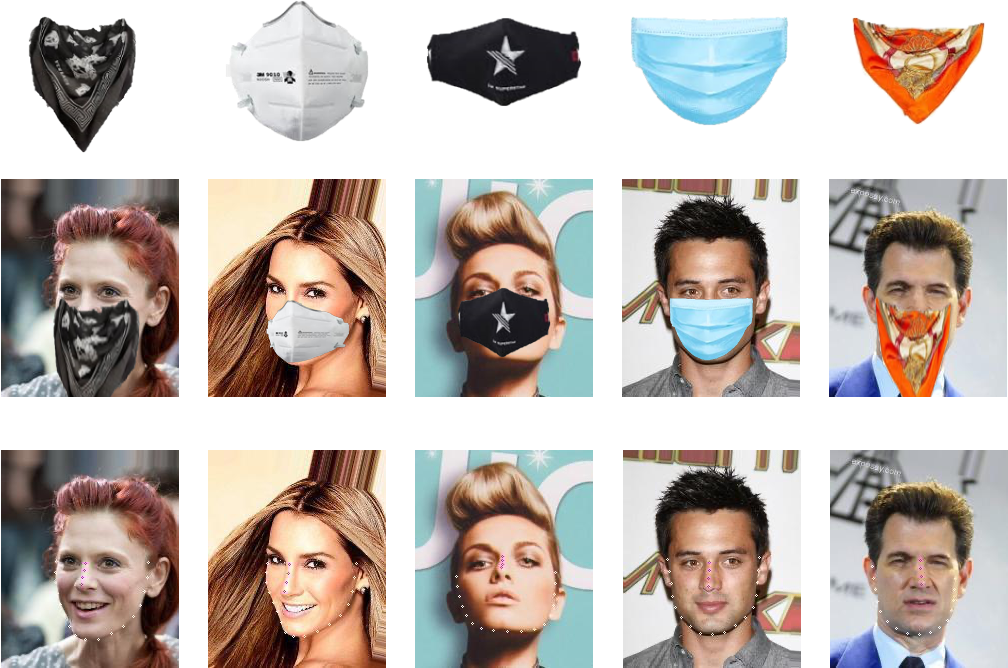}
\end{center}
\vspace{-0.3cm}
  \caption{\textbf{Illustration of how Masked-Faces dataset is synthesized.} Top row shows the five types of masks used. Last row shows the output of face detection with 21 facial landmarks. Middle row shows the masks aligned and placed with respect to the landmarks.}
\label{fig:Mask}
\end{figure}

\label{sec:experiments}
\section{Experiments and the Masked-Faces Dataset}
\label{sec:experiments}
In this section, we first introduce a new synthetic dataset we generated,  Masked-Faces dataset. We then describe the implementation details. Finally, we compare our method with other SOTA methods in terms of the peak signal-to-noise ratio (PSNR), the structural similarity index measure (SSIM) and the $\ell_1$ loss.

\begin{figure}[ht!]
\centering
\begin{subfigure}{0.24\linewidth}
        \includegraphics[width=1.0\linewidth]{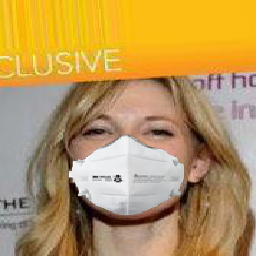}\\
        \includegraphics[width=1.0\linewidth]{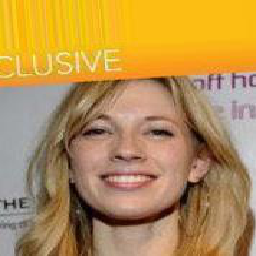}\\
        \includegraphics[width=1.0\linewidth]{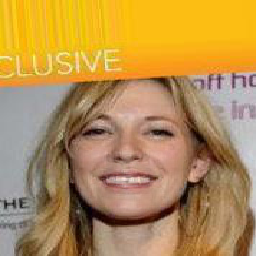}\\
        \includegraphics[width=1.0\linewidth]{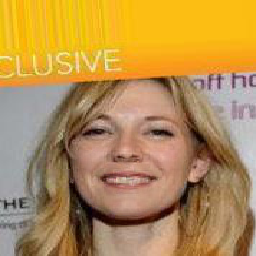}
    \caption{}
    \label{fig:ab2_a}
\end{subfigure}
\begin{subfigure}{0.24\linewidth}
        \includegraphics[width=1.0\linewidth]{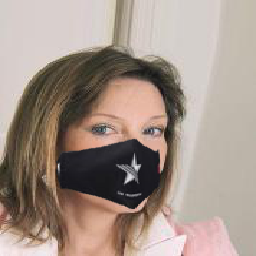}\\
        \includegraphics[width=1.0\linewidth]{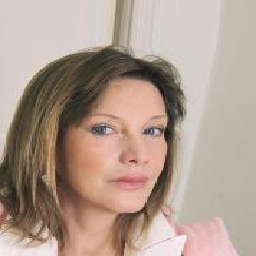}\\
        \includegraphics[width=1.0\linewidth]{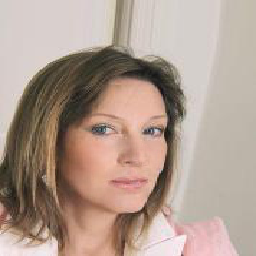}\\
        \includegraphics[width=1.0\linewidth]{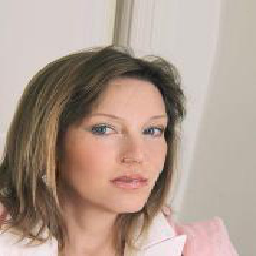}
    \caption{}
    \label{fig:ab2_b}
\end{subfigure}
\begin{subfigure}{0.24\linewidth}
        \includegraphics[width=1.0\linewidth]{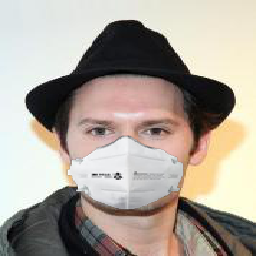}\\
        \includegraphics[width=1.0\linewidth]{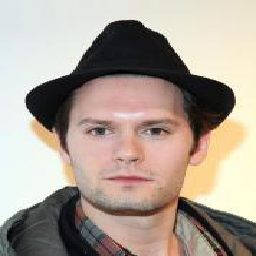}\\
        \includegraphics[width=1.0\linewidth]{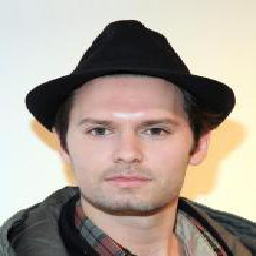}\\
        \includegraphics[width=1.0\linewidth]{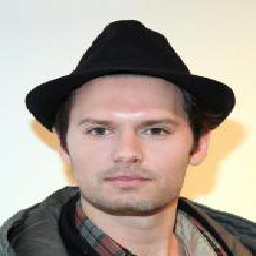}
    \caption{}
    \label{fig:ab2_c}
\end{subfigure}
\begin{subfigure}{0.24\linewidth}
        \includegraphics[width=1.0\linewidth]{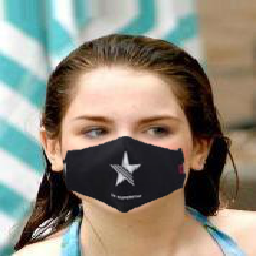}\\
        \includegraphics[width=1.0\linewidth]{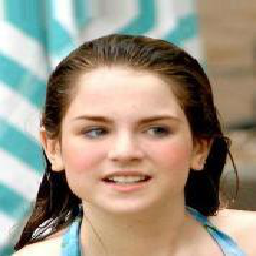}\\
        \includegraphics[width=1.0\linewidth]{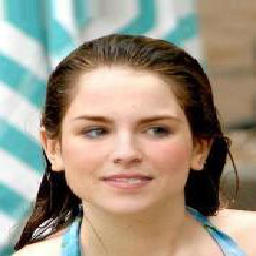}\\
        \includegraphics[width=1.0\linewidth]{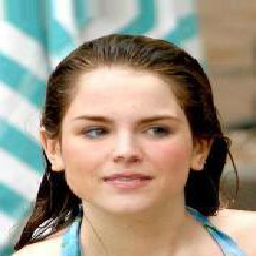}
    \caption{}
    \label{fig:ab2_d}
\end{subfigure}
\vspace{-0.2cm}
\caption{\textbf{Qualitative comparison of models with and without CSAM.} First row: input masked images; Second row: ground-truth images; Third row: results from the network with CSAM incorporated; Forth row: results from the network without CSAM (best viewed when zoomed-in).}
\label{fig:abs_csam}
\end{figure}

\begin{figure*}[th!]
\centering
\scalebox{0.9}{
\begin{subfigure}{0.13\linewidth}
        \includegraphics[width=1.0\linewidth]{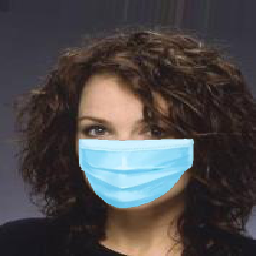}\\
        \includegraphics[width=1.0\linewidth]{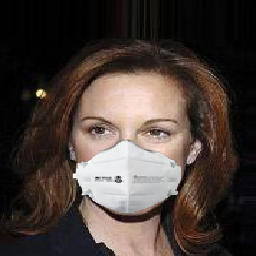}\\
        \includegraphics[width=1.0\linewidth]{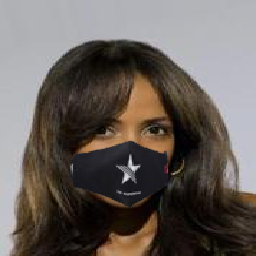}\\
        \includegraphics[width=1.0\linewidth]{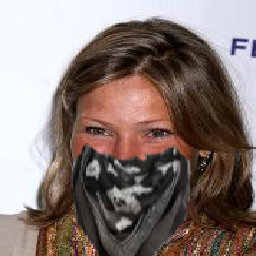}\\
        \includegraphics[width=1.0\linewidth]{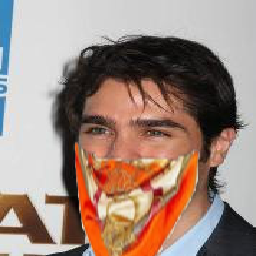}
    \caption{Input}
    \label{fig:a}
\end{subfigure}
\begin{subfigure}{0.13\linewidth}
        \includegraphics[width=1.0\linewidth]{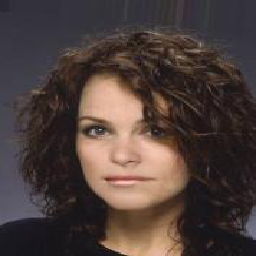}\\
        \includegraphics[width=1.0\linewidth]{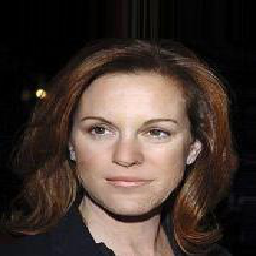}\\
        \includegraphics[width=1.0\linewidth]{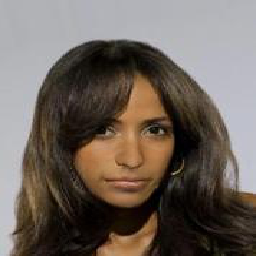}\\
        \includegraphics[width=1.0\linewidth]{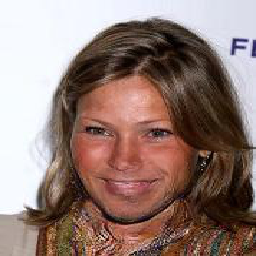}\\
        \includegraphics[width=1.0\linewidth]{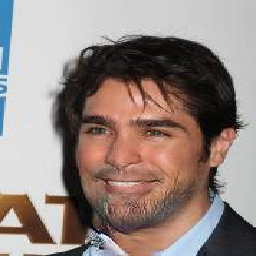}
    \caption{GatedConv\cite{yu2019free}}
    \label{fig:b}
\end{subfigure}
\begin{subfigure}{0.13\linewidth}
        \includegraphics[width=1.0\linewidth]{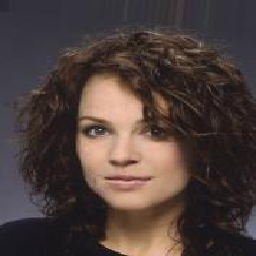}\\
        \includegraphics[width=1.0\linewidth]{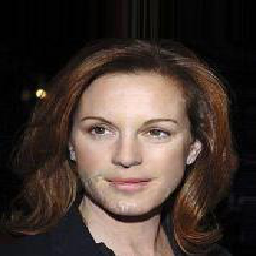}\\
        \includegraphics[width=1.0\linewidth]{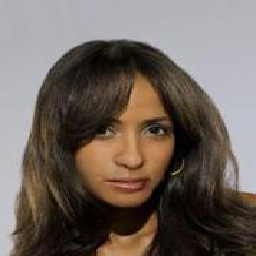}\\
        \includegraphics[width=1.0\linewidth]{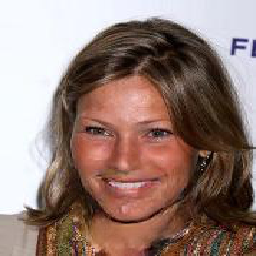}\\
        \includegraphics[width=1.0\linewidth]{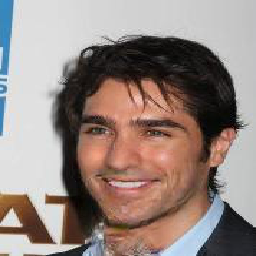}
    \caption{CSA\cite{liu2019coherent}}
    \label{fig:c}
\end{subfigure}
\begin{subfigure}{0.13\linewidth}

        \includegraphics[width=1.0\linewidth]{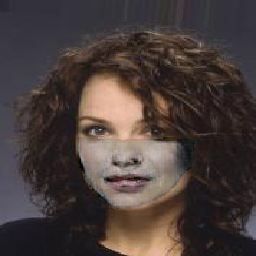}\\
        \includegraphics[width=1.0\linewidth]{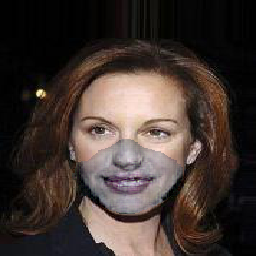}\\
        \includegraphics[width=1.0\linewidth]{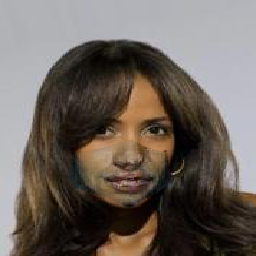}\\
        \includegraphics[width=1.0\linewidth]{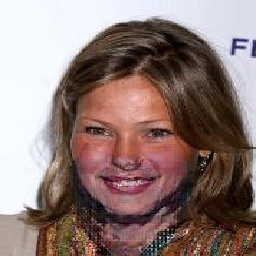}\\
        \includegraphics[width=1.0\linewidth]{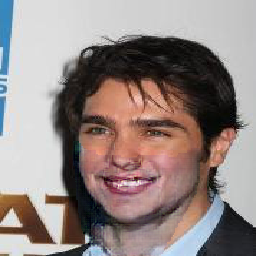}
    \caption{RFR\cite{li2020recurrent}}
    \label{fig:d}
\end{subfigure}
\begin{subfigure}{0.13\linewidth}
        \includegraphics[width=1.0\linewidth]{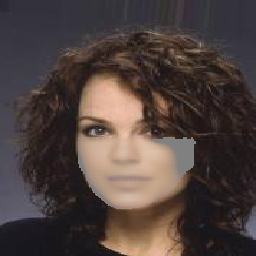}\\
        \includegraphics[width=1.0\linewidth]{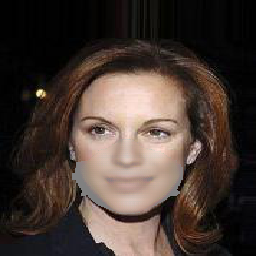}\\
        \includegraphics[width=1.0\linewidth]{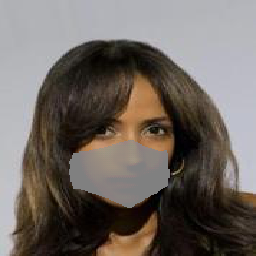}\\
        \includegraphics[width=1.0\linewidth]{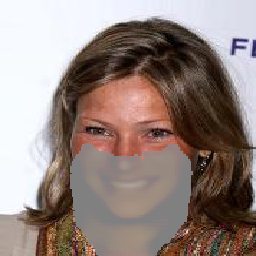}\\
        \includegraphics[width=1.0\linewidth]{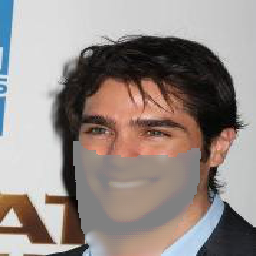}
    \caption{RN\cite{yu2020region}}
    \label{fig:e}
\end{subfigure}
\begin{subfigure}{0.13\linewidth}
        \includegraphics[width=1.0\linewidth]{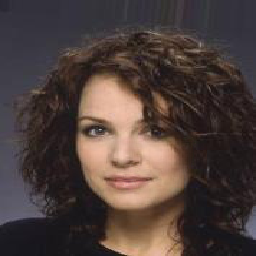}\\
        \includegraphics[width=1.0\linewidth]{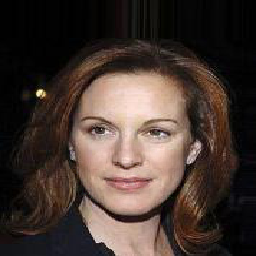}\\
        \includegraphics[width=1.0\linewidth]{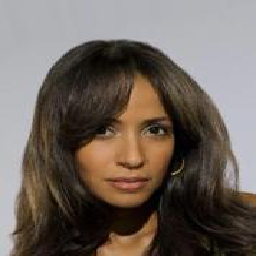}\\
        \includegraphics[width=1.0\linewidth]{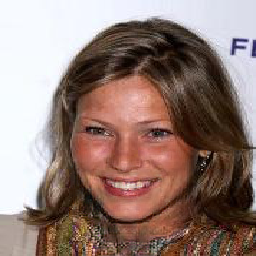}\\
        \includegraphics[width=1.0\linewidth]{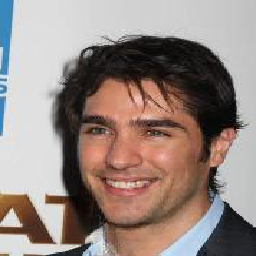}
        \caption{Ours}
        \label{fig:f}
\end{subfigure}
\begin{subfigure}{0.13\linewidth}
        \includegraphics[width=1.0\linewidth]{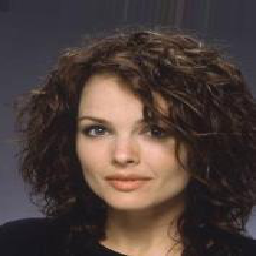}\\
        \includegraphics[width=1.0\linewidth]{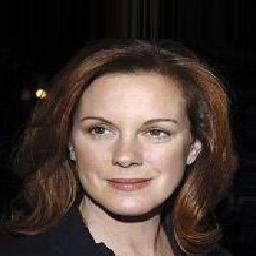}\\
        \includegraphics[width=1.0\linewidth]{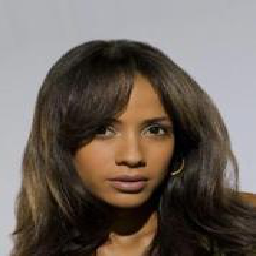}\\
        \includegraphics[width=1.0\linewidth]{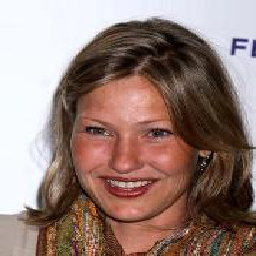}\\
        \includegraphics[width=1.0\linewidth]{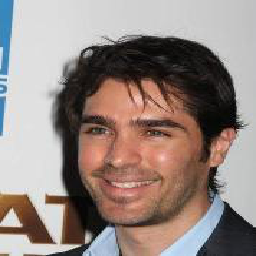}
        \caption{Ground-truth}
        \label{fig:g}
\end{subfigure}}
\caption{\textbf{Qualitative comparison.} Different approaches are compared on the Masked-Faces dataset.}
\label{fig:exp1}
\end{figure*}

\subsection{Masked-Faces Dataset Construction}
\label{sec:Dataset}
We constructed a new Masked-Faces dataset, from the CelebA dataset~\cite{liu2015faceattributes}, by putting on face masks on face images to simulate real-world masked faces.~The CelebA dataset~\cite{liu2015faceattributes} contains $202,599$ face images with large pose variance. We first cropped the images to the same size of $256\times256$ pixels. We then used dlib's face detector~\cite{dlib09} to find locations of faces and get 21 facial landmarks, including 4 landmarks for nose bridge and 17 landmarks for chin, as shown in the third row of \cref{fig:Mask}.  We employ five different types of masks, having different colors and shapes, as shown in the first row of \cref{fig:Mask}. There are surgical masks, regular cloth masks, N95-type masks, and scarves. Each face image is masked with a randomly chosen mask to get the masked face image. We reshape (i.e., rotate and resize) the mask image according to the facial landmarks. The height of the mask is resized according to the length from nose to the bottom of chin, and the width of the mask is resized based on the landmarks of chin and nose. Due to high variance of face orientations, not all 21 landmarks could be found for some face images in the CelebA dataset. Therefore, there are a total of 196,999 masked face images. 

\subsection{Implementation Details}
The CelebA dataset~\cite{liu2015faceattributes} provides public training, evaluation and test sets. We use the same training, validation and test splits for our Masked-Faces dataset. Our model is trained on the training set (158,003 images), and evaluated on the validation set (19,026 images) of Masked-Faces dataset. Data augmentation, such as rotation, is adopted during training. We use the Adam optimizer to train the model with $\beta_1=0.5$ and $\beta_2 = 0.999$. The learning scheduler has a base learning rate of $2\times10^{-4}$, and decreases the learning rate by $2\times10^{-6}$ for each epoch after 20 epochs. The individual loss weights are set as $\lambda_r=1$, $\lambda_p=0.1$, $\lambda_s=250$, $\lambda_{adv}=0.1$. Our model is trained on four NVIDIA RTX6000 GPUs (24GB) with a batch size of 16.

\subsection{Performance Evaluation}
We compare our method with four state-of-the-art methods, namely GatedConv~\cite{yu2019free}, CSA~\cite{liu2019coherent}, RFR~\cite{li2020recurrent} and RN~\cite{yu2020region}, on our Masked-Faces dataset.  \vspace{-0.3cm}

\paragraph{Quantitative Comparison:}
We test all the methods on the validation data ($19,026$ images in total) of the Masked-Faces dataset. For quantitative comparison, we employ the metrics of SSIM~\cite{wang2004image}, PSNR and $\ell_1$ loss. We calculate the SSIM~\cite{wang2004image} between the recovered image and the ground truth. The results are provided in \cref{table:1}. As can be seen, our proposed method outperforms all the other works in terms of the SSIM~\cite{wang2004image}, PSNR and $\ell_1$ loss. 


\begin{table}[h!]
\centering
\scalebox{1}{
\begin{tabular}{ @{}l c c r@{}}
 \hline
 & SSIM~\cite{wang2004image}$^+$  &  PSNR$^+$ & $\ell_1^-$ loss\\
 \hline
 GatedConv\cite{yu2019free}   & 0.9428    &29.6119  & 0.0096 \\
 CSA\cite{liu2019coherent}   & 0.9394  & 28.7021   & 0.0104\\
 RFR\cite{li2020recurrent}   & 0.9260    &27.0240   & 0.0116\\
 RN\cite{yu2020region}   & 0.8803  & 24.4053  & 0.0372 \\
 Ours & \textbf{0.9511} & \textbf{30.3885} & \textbf{0.0076}\\
 \hline
\end{tabular}}
\caption{\textbf{Quantitative results on the Masked-Faces dataset.} Binary masks are from a segmentation network. $^+$Higher is better. $^-$Lower is better.}
\label{table:1}
\end{table}
\vspace{-0.3cm}
\paragraph{Qualitative Comparison:}
Example output images obtained with different methods are shown in \cref{fig:exp1} for qualitative comparison. As can be seen, the GatedConv~\cite{yu2019free} recovers masked areas with blurry content. CSA~\cite{liu2019coherent} performs better due to its coherent semantic attention module, but there are imperfections in its predictions, which present distorted structure and inconsistent color (\eg the neck areas for the last two rows of the figure). RFR~\cite{li2020recurrent} has the issue of color distortion and line discontinuities. RN~\cite{yu2020region} cannot fully remove masks and cannot recover the masked region with consistent color. Compared to other four methods, our proposed model can restore the masked areas with more consistent structure and color coherence in the masked regions.


\begin{figure*}[ht]
\centering
\begin{subfigure}{0.12\linewidth}
        \includegraphics[width=1.0\linewidth]{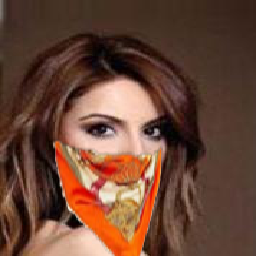}\\
        \includegraphics[width=1.0\linewidth]{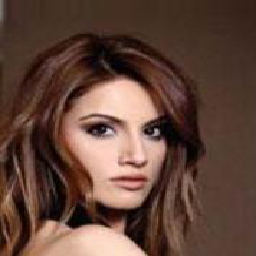}\\
        \includegraphics[width=1.0\linewidth]{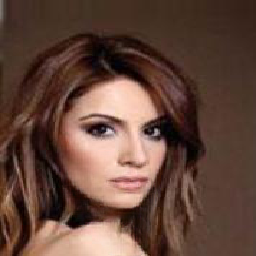}\\
        \includegraphics[width=1.0\linewidth]{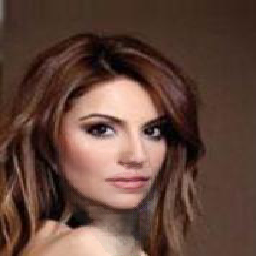}
    \caption{}
    \label{fig:ab_a}
\end{subfigure}
\begin{subfigure}{0.12\linewidth}
        \includegraphics[width=1.0\linewidth]{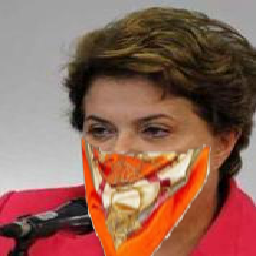}\\
        \includegraphics[width=1.0\linewidth]{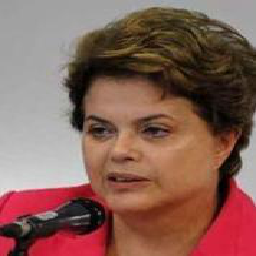}\\
        \includegraphics[width=1.0\linewidth]{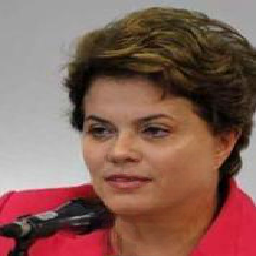}\\
        \includegraphics[width=1.0\linewidth]{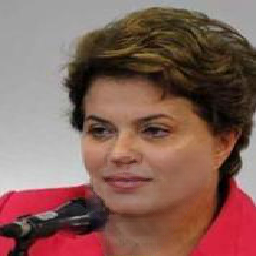}
    \caption{}
    \label{fig:ab_b}
\end{subfigure}
\begin{subfigure}{0.12\linewidth}
        \includegraphics[width=1.0\linewidth]{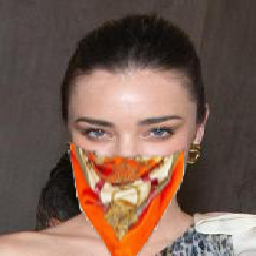}\\
        \includegraphics[width=1.0\linewidth]{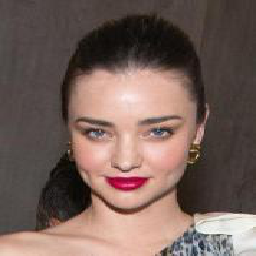}\\
        \includegraphics[width=1.0\linewidth]{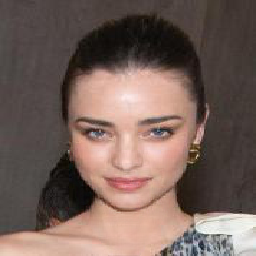}\\
        \includegraphics[width=1.0\linewidth]{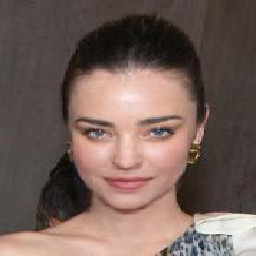}
    \caption{}
    \label{fig:ab_c}
\end{subfigure}
\begin{subfigure}{0.12\linewidth}
        \includegraphics[width=1.0\linewidth]{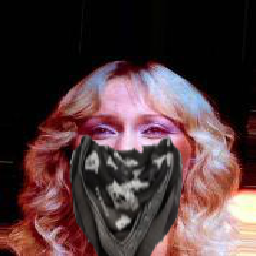}\\
        \includegraphics[width=1.0\linewidth]{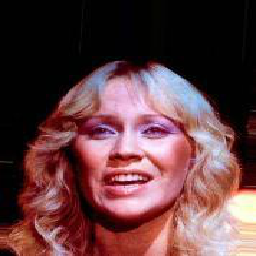}\\
        \includegraphics[width=1.0\linewidth]{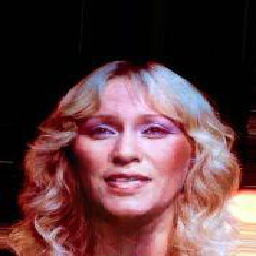}\\
        \includegraphics[width=1.0\linewidth]{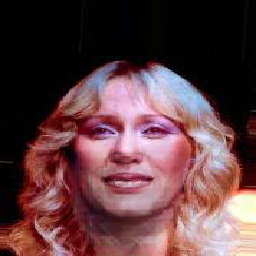}
    \caption{}
    \label{fig:ab_d}
\end{subfigure}
\begin{subfigure}{0.12\linewidth}
        \includegraphics[width=1.0\linewidth]{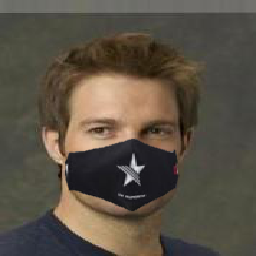}\\
        \includegraphics[width=1.0\linewidth]{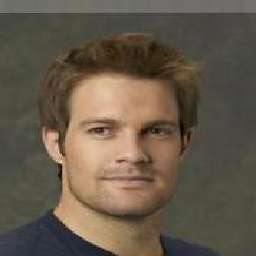}\\
        \includegraphics[width=1.0\linewidth]{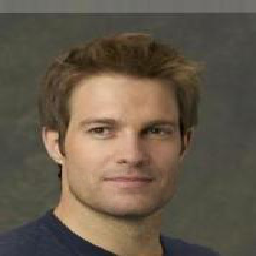}\\
        \includegraphics[width=1.0\linewidth]{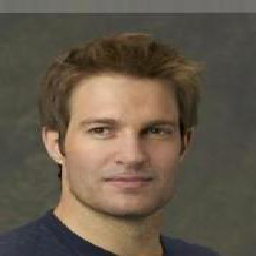}
    \caption{}
    \label{fig:ab_e}
\end{subfigure}
\begin{subfigure}{0.12\linewidth}
        \includegraphics[width=1.0\linewidth]{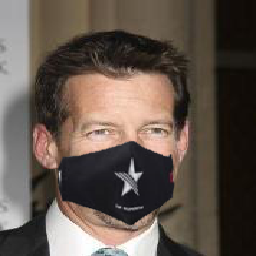}\\
        \includegraphics[width=1.0\linewidth]{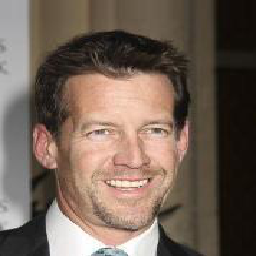}\\
        \includegraphics[width=1.0\linewidth]{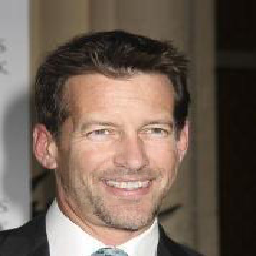}\\
        \includegraphics[width=1.0\linewidth]{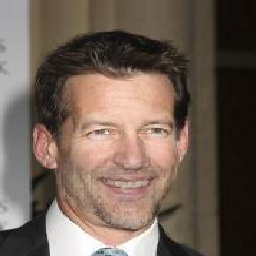}
        \caption{}
        \label{fig:ab_f}
\end{subfigure}
\vspace{-0.2cm}
\caption{\textbf{Qualitative comparison of local area supervision and the supervision over the whole image}. First and second rows are the input masked image and ground-truth images, respectively. The third and fourth rows show the results from local supervision and from full supervision, respectively. }
\label{fig:local_area}
\vspace{-0.3cm}
\end{figure*}

\subsection{Face Recognition Results}
\label{sec:Rcg}
\vspace{-0.1cm}
For this experiment, we employ two different face recognition networks, namely MobileFaceNet~\cite{chen2018mobilefacenets} and ArcFace~\cite{deng2019arcface}, to show (i) the degradation of face recognition performance when masked face images are used, and (ii) the improvement after face inpainting.~Table~\ref{table:FR} reveals two key observations.~First, face masks deteriorate the face recognition performance considerably (by 7.85\% on average for 2 networks), whereas using the inpainted images, from all methods, results in higher face recognition accuracy, with images inpainted by our approach resulting in the highest accuracy. Second, the accuracy comparison across all inpainting approaches show that our approach has a better identity-preserving ability to facilitate face recognition, which can be explained by our method's ability to recover the masked region with better-quality details and higher similarity to ground truth.

\begin{table}[h!]
\vspace{-0.2cm}
\centering
\scalebox{0.88}{
\begin{tabular}{ @{}l | c r@{}}
 \hline
 Input         & MobileFaceNet~\cite{chen2018mobilefacenets} &   ArcFace~\cite{deng2019arcface} \\
 \hline
  Grnd Truth Img.            & 77.61         &  76.44    \\
 Masked Image  & 68.31         &  70.04    \\ \hline
 GatedConv~\cite{yu2019free}     & 71.72         &  72.08    \\
 CSA~\cite{liu2019coherent}           & 72.13         &  72.54    \\
 RFR~\cite{li2020recurrent}           & 71.34         &  73.71    \\
 RN~\cite{yu2020region}            & 72.93         &  72.37    \\
 Ours          & \textbf{75.67}         &  \textbf{73.87}    \\

 \hline
\end{tabular}
}

\caption{\small{\textbf{Face Recognition Accuracy (\%).} Among the inpainted images, the ones obtained with our method result in the highest face recognition accuracy.}}
\label{table:FR}
\end{table}

\subsection{Ablation Study}
\paragraph{Effect of local area supervision.}
Instead of supervising the generation of whole images, our network focuses only on recovering the masked region, which has lower pixel variance compared to the whole image. We conduct further experiments to verify our conjecture that local error supervision is better than full image supervision. We train the model in two ways: (i) one focusing on the direct output from the inpainting network, $I_r$ and (ii) one focusing on the synthetic output, $I_{syn}$. The results are provided in \cref{table:lossAblation}, showing that supervision only over the masked region is beneficial for learning in terms of SSIM~\cite{wang2004image}, PSNR and $\ell_1$ loss. 
\begin{table}[h]
\vspace{-0.1cm}
\centering
\scalebox{1}{
\begin{tabular}{ @{}l c c r @{}}
 \hline
 & SSIM~\cite{wang2004image}$^+$  &  PSNR$^+$  & $\ell_1^-$ loss\\
 \hline
 local supervision   & \textbf{0.9511} & \textbf{30.3885} & \textbf{0.0076}\\
 full supervision   & 0.9441  & 29.6932  & 0.0088\\
 \hline
\end{tabular}}
\vspace{-0.2cm}
\caption{The comparison of supervision over masked region and the whole image. $^+$Higher is better. $^-$Lower is better.}
\label{table:lossAblation}
\vspace{-0.2cm}
\end{table}

The qualitative comparison between two ways of supervision is provided in \cref{fig:local_area}. The results from the local supervision have less color diffusion and structure distortion. However, for the results from full supervision, the recovered pixels have larger color discrepancy with the surrounding unmasked pixels when comparing the third and forth rows in \cref{fig:ab_a}, \cref{fig:ab_b}, \cref{fig:ab_d} and \cref{fig:ab_f}. In addition, when the third and forth rows of \cref{fig:ab_a} and \cref{fig:ab_d} are compared, it can be seen that the local supervision strategy can help the network generate more plausible and clear texture, such as the texture of nose and mouth.
\vspace{-0.2cm}
\paragraph{Effect of CSAM.}
We performed an ablation study to test the effectiveness of CSAM, by keeping the four dilated convolutional layers, and replacing the CSAM module with a conventional $3 \times 3$ convolutional layer. The numerical comparison is shown in \cref{table:csamAblation}. As shown in \cref{fig:abs_csam}, the use of CSAM helps the network generate more reasonable content. The details of nostril is better recovered when CSAM is incorporated. The results obtained with M-CSAM are provided in the Supplementary material.  
\begin{table}[h]
\vspace{-0.1cm}
\centering
\scalebox{0.95}{
\begin{tabular}{ @{}l c c r @{}}
 \hline
 & SSIM~\cite{wang2004image}$^+$  &  PSNR$^+$  & $\ell_1^-$ loss\\
 \hline
w/ CSAM   & \textbf{0.9511} & \textbf{30.3885} & \textbf{0.0076}\\
w/o CSAM   & 0.9495  & 30.1338  & 0.0077\\
 \hline
\end{tabular}}
\vspace{-0.2cm}
\caption{The effect of CSAM. $^+$Higher is better. $^-$Lower is better.}
\label{table:csamAblation}
\vspace{-0.2cm}
\end{table}

\label{sec:conclusion}
\section{Discussion and Conclusion}

With the COVID-19 pandemic, face masks have become ubiquitous in our lives. Removing face masks from images/videos can be desirable for varies reasons including face recognition, and image/video editing and enhancement purposes. We have presented a face inpainting method, based on a generative network with our proposed Multi-scale Channel-Spatial Attention Module (M-CSAM), to learn the inter- and intra-channel correlations. In addition, we have introduced a method to enforce supervision on masked regions instead of the entire face image. In addition, we have presented a new dataset, Masked-Faces, which we have synthesized from the CelebA dataset by using five different types of masks. Experiments performed on the Masked-Faces dataset have shown that our proposed approach can outperform four state-of-the-art methods in terms of structural similarity index measure and PSNR metrics, as well as qualitative comparison. 

Our proposed architecture contains 44.7M parameters, which is significantly less than the number of parameters of CSA (132.1M), but more than those of RFR (31.2M), RN (11.6M) and GatedConv (4.1M). Thus, one limitation is that the SOTA performance achieved by the proposed method comes with a larger network. In addition, we synthesized masked face images to have a large dataset with varying mask shapes and sizes. These synthesized images may not always substitute for more naturalistic images.


{\small
\bibliographystyle{cvpr22main}
\bibliography{cvpr22main}
}

\end{document}